# UROLOGIC ROBOTS AND FUTURE DIRECTIONS

Pierre Mozer MD[1], Jocelyne Troccaz PhD[2], Dan Stoianovici PhD[1]

[1]Urology Robotics, Johns Hopkins University, Baltimore, MD
http://urobotics.urology.jhu.edu/

[2]TIMC, Grenoble, France

Corresponding author:
**Dan Stoianovici, PhD**
JHBMC, URobotics Lab, MFL-W115
5200 Eastern Ave.
Baltimore, MD 21224, USA
Tel:    410-550-1980
Elm:    dss@jhu.edu

**Abstract** :

Robot-assisted laparoscopic surgery in urology has gained immense popularity with the Da Vinci system but a lot of research teams are working on new robots. The final goal of robots is to allow safer and more homogeneous outcomes with less variability of surgeon performance, as well as new tools to perform tasks based on medical transcutaneous imaging, in a less invasive way, at lower costs. The purpose of this paper is to review current urologic robots and present future developments directions.
Based on the Computer Aided Surgery concepts, future systems are expected to advance in two directions: improvements of remote manipulation robots and developments of image-guided robots. For each direction, current research trends are presented with a clinical point of view. It is expected that improvements for remote system could be augmented reality, haptic feed back, size reduction and development of new tools for NOTES surgery. The paradigm of image-guided robots is close to a clinical availability and the most advanced robots are presented with end-user technical assessments. It is also notable that the potential of robots lies much further ahead than the accomplishments of the daVinci system. The integration of imaging with robotics holds a substantial promise, because this can accomplish tasks otherwise impossible. Image guided robots have the potential to offer a paradigm shift.

# I. Introduction

A robot is a mechanical device controlled by a computer. Medical robots have been classified in several ways. Three types were distinguished from an operational point of view [1]: remote controlled, synergistic [2], and automated or semi-automated robots. In the first two types, the physician has direct real-time control of the robotic instrument either from a console, or by handling the instrument itself. For the later class, the physician does not have to continuously control the motion of the robot, but rather define its task and monitor the execution.

Robots in different categories are significantly dissimilar from the technical point of view, having other design requirements. It is commonly the case that directly controlled robots have less precision requirements because the motion is compensated by the physician but have additional complexity for implementing the direct control of the physician. Image-guided robots do not normally need a surgeon console, but need to be more accurate and precise to operate without human compensation.

This article gives a short presentation of the achievements and developments in the field, a few key concepts of robotics and medical robotics, and several examples of these technologies commercial and under development.

# II. Urology Robots in Current Clinical Use

The daVinci® Surgical System (Intuitive Surgical, Inc., Sunnyvale, CA) platform is the only robotic system use in common practice with more than 800 robots installed worldwide. The robot is a surgeon-driven system with 3 ou 4 arms allowing endowrist capabilities and a 3D visualization of the surgical field. Even though several drawbacks have been echoed about its functionality and possible improvements, this system popularized the concept and instrumentation of robotic surgery.

In large majority the robots are used for robotic assisted laparoscopic radical prostatectomy (RALP) [3]. Even if the review of published literature on RALP and open radical prostatectomy (ORP) is currently insufficient to favour one surgical technique, it seems that short-term outcomes of RALP achieve equivalence to open surgery with regards to complications and functional results [4]. Applications to bladder cancer, renal cancer, uretero-pelvic junction obstruction, and pelvic prolapse have also been explored [5]. The main technical improvement since the first release of the system was the addition of a fourth robotic arm, yet other features especially with respect to improved sensory feedback could significantly improve its performance and surgeon acceptance.

# III. Future

Developments aim systems with decreased learning curves that would allow for safer and more homogeneous outcomes with less variability of surgeon performance, as well as new tools to perform more autonomous tasks in a less invasive way at lower costs. Conceptually, robotic developments are an integral part of the Computer Aided Surgery (CAS) paradigm [6]. This integrates preoperative planning, intraoperative guidance, robotic assistance, and postoperative verification and follow-up. Augmented reality is a part of this concept including image fusion from various imaging modalities, such as preoperative CT with laparoscopic images [7] or CT and ultrasound images [8].

Based on the CAS concepts, future systems are expected to advance in the following two directions: Improvements of remote manipulation robots for surgery, developments of image-guided robots for interventions, and possibly combining the two categories.

## A. Remote Manipulation Robots

Current surgical robotic research shows a trend of size reduction compared to the daVinci system. For example, the NeuroArm (University of Calgary, Canada) proceeds with the development of a remotely controlled bilateral arm robot for neurosurgical operations. Part of the scope is to reduce its size to where the robot could be brought in the bore of an MRI scanner. Even though this is not yet possible, their current version is substantially smaller than the daVinci, and has additional features such as force feedback [9]. Another example is the VikY system [10], which is a very compact robot allowing to move a laparoscopic camera. Technical works to hold surgical tools on this platform are ongoing (Figure 1).

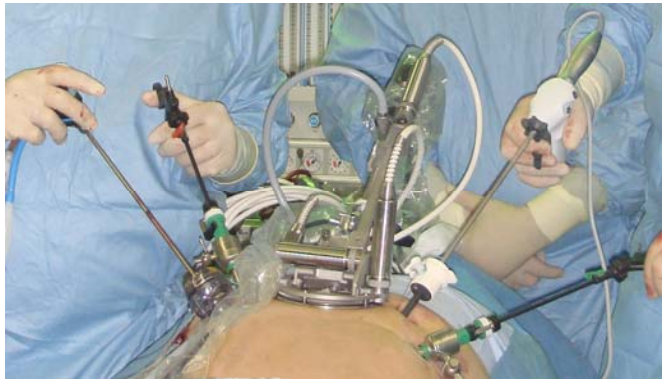

Figure 1: VikY Robot

A common concern with the daVinci robot is the lack of haptic feedback. Several teams are pursuing additions to the existing system for augmenting sensory feedback [11] and with modified trocar instruments for allowing the measurement of manipulation forces [12].

A novel approach is pursuing the development of tools to be deployed in the peritoneal cavity and controlled externally with magnetic fields for reducing the number of transabdominal trocars and for increasing the range of motion and accessibility [13].

The development of natural orifice translumenal endoscopic surgery (NOTES) is potentially the next paradigm shift in minimally invasive surgery. The concept is to access to the peritoneal cavity without passing through the anterior abdominal wall. The first clinical case, performed in 2007, was a cholecystectomy in a woman via a transvaginal approach [14]. Nevertheless, NOTES procedures are performed using modified endoscopic tools with significant constraints and new tools are necessary to allow the surgeon to better visualize and dexterously manipulate within the surgical environment. A two-armed dexterous miniature robot with stereoscopic vision capabilities is under development [15].

## B. Direct Image-Guided Robots

Traditionally, image-guidance and navigation of instruments has been performed manually based on pre-acquired images with the use of spatial localizers such as optical [16] and magnetic trackers [17]. However, robots have the potential to improve the precision, accuracy, and reliability of performance in image-guidance interventions because the tasks are done in a full digital way, from image to instrument manipulation.

Robots for interventions with needles or other slender probes or instruments can be connected to an imaging modality (CT, MRI, ultrasound, fluoroscopy, etc.). Targets and paths are defined in the image based on planning algorithms, and the robot aligns and may insert the needle accordingly. The true potential of needle delivery mechanisms relies on their ability to operate with, be guided by, and use feedback from medical imaging equipment.

Moreover, robots can do complex movements, impossible to perform by a human to limit tissue and needle deformations during the insertion. Indeed, decreasing the force of needle insertion has been proposed with special movements as for example a rotating needle for increasing the accuracy of reaching a target [18]. Another option as suggested by Podder and al. [19] is to insert multiple needles simultaneously for prostate therapies to reduce operative time.

**Image input:**

Image-guided robots have stringent requirements for imager compatibility, precision, sterility, safety, as well as size and ergonomics [6]. A robot's compatibility with a medical imager refers to the capability of the robot to safely operate within the confined space of the imager while performing its clinical function, without interfering with the functionality of the imager [20].

The current research trend is to embed the robot with the imager (CT, MRI, ultrasound, fluoroscopy, etc.) for re-imaging during the intervention, for relocalization treatment planning updates and quality control. We term these procedures Direct Image-Guided Interventions (DIGI). The performance of DIGI interventions is not new, in fact the routine TRUS biopsy is done under direct guidance, however the new term is essential for distinguishing this important class of Image Guided Intervention (IGI) from navigation based on pre-acquired imaging data.

Among all types of imagers, the MRI is the most demanding and the development of MRI compatible robots is a very challenging engineering task [21]. But, this also makes MRI compatible multi-imager compatible, if care is taken for the selection of radiolucent materials for the components in immediate proximity of the imaging site [20]. Due to the strong requirements needed to build a MRI compatible robot, the following description of many robots under development is presented with respect to their capabilities of operation leading up to use with the MRI.

**Robots using X-Ray based and Ultrasound Guidance:**

From historic point of view the first systems were robots with image-guided capabilities. Davies developed a robot for prostatectomies, called Probot [22], based on an industrial Unimate Puma robot constrained within a frame for safety consideration. The robot was guided by transrectal ultrasound images and it was the first robotic device used to remove tissue from a patient when it underwent its first clinical trial in March of 1991.

A few years later, the URobot system was developed in Singapore by Ng and al. The robot was designed to performed a trans-urethral and trans-perinal access to the prostate for laser resection in 2001 [23] or brachytherapy [24] respectively.

Professor Brian Davies has also reported the development of a simple robot that performs similar to the brachytherapy template [25]. Rotation about the axis of the needle is added in order to reduce needle deflections. The system uses 2D TRUS guidance and the report describes successful preclinical testing.

In the Robarts Research Institute (London, Canada) [26] and in the Nanyang Technological University (Singapore) [27], three dimensional (3D) reconstruction from a regular 2D TRUS probe has been investigated by sweeping the probe about its axis. This was integrated with a robot in a system for prostate brachytherapy or biopsy. Mockup tests demonstrated a precision on the order of one millimeter and a clinical study for biopsy is ongoing in Singapore.

Our URobotics laboratory at Johns Hopkins has also developed several versions of a CT-guided robot and performed numerous clinical tests for urology applications [28]. Recently, the AcuBot robot was instrumented with a new end-effecter, the Revolving Needle Driver (RND). The RND is a fully actuated driver for needle insertion, spinning, release, and force measurement (Figure 2). The driver supports the needle from its head, and provides an additional needle support guide in close proximity of the skin entry point. This is similar to holding the needle with two finger-like grippers, one from its head and one from its barrel next to the skin. The top one pushes the needle in and out, while the lower holds the guide to support the direction of the needle as close as possible to the skin. Both grippers can simultaneously release the needle automatically. Finally, the new driver is also equipped with a set of force sensors to measure the interaction of the nozzle with the patient and the force of needle insertion [29].

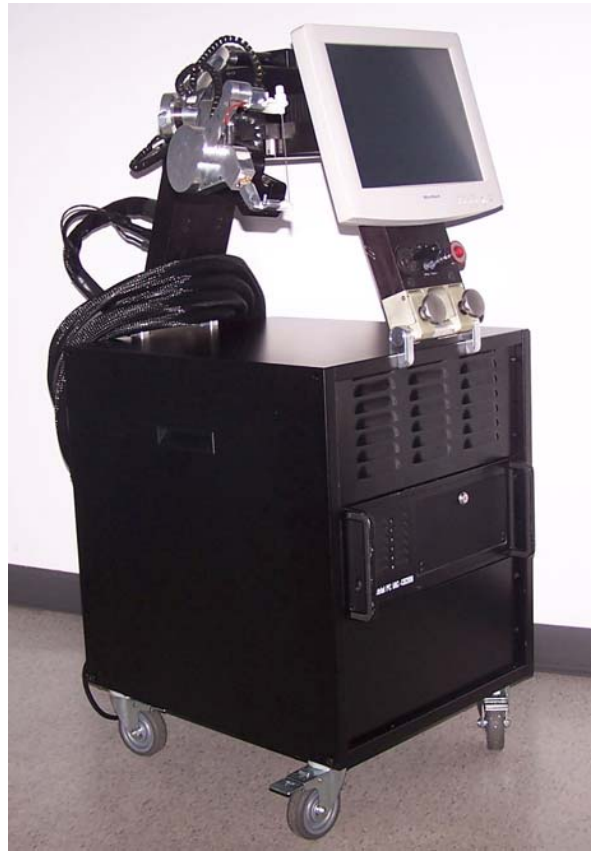

Figure 2: Revolving Needle Driver on the AcuBot robot

**MRI Compatible Robots:**

The earliest work for MRI guided prostate intervention robots was performed at the Brigham and Women's Hospital (BWH), Boston MA in collaboration with AIST-MITI, Japan [30, 31]. A robotic intervention assistant was constructed for open MRI to provide a guide for needles and probes [32, 33]. The system assists the physician by positioning a needle guide for manual needle intervention. Applications included prostate biopsy and brachytherapy [34] but resolution and functional imaging are limited by the low magnetic field of the open MR.

The Institute for Medical Engineering and Biophysics (IMB), Karlsruhe, Germany reported several versions of a robotic system for breast lesion biopsy and therapy under MR guidance [35, 36]. Currently, the development is done by Innomedic (Germany) and their last version used a cylinder for driving an end-effector axis [37]. The robot orients the needle about the axial-sagittal planes for interventions targeting abdominal organs. However, a group from Frankfurt, Germany has recently used the Innomotion system for targeting the prostate [38]. The limitations of the robot restricted the access to the transgluteal path (prone patient with needle pointing down) for which the needle path is much deeper than normal (~14 cm reported in the cadever experiment) [20]. A 15Ga needle was needed to prevent deflections. Manual needle insertion was performed through the guide after retracting the table from the scanner. Even though the Innomedic system is not FDA approved and its designed application range does not include the prostate, it is approved for clinical use in Europe and is a commercial DIGI robot.

Our group at Johns Hopkins has also developed an MRI compatible robot for prostate access [39]. MrBot, was constructed to be multi-imager compatible, which includes compatibility with all classes of medical imaging equipment (ultrasound, X-Ray, and MR based imagers) [40]. All robotic components are constructed of nonmagnetic and dielectric materials. To overcome MRI incompatibilities a new type of motor, PneuStep [9], was purposely designed. The robot presents 6 degrees of freedom (DOF), 5 for positioning and orienting the injector, and one for setting the depth of needle insertion. Various needle drivers can be mounted in the robot for performing various needle interventions. The first driver was developed for fully automated low dose (seed) brachytherapy [41] (Figure 3).

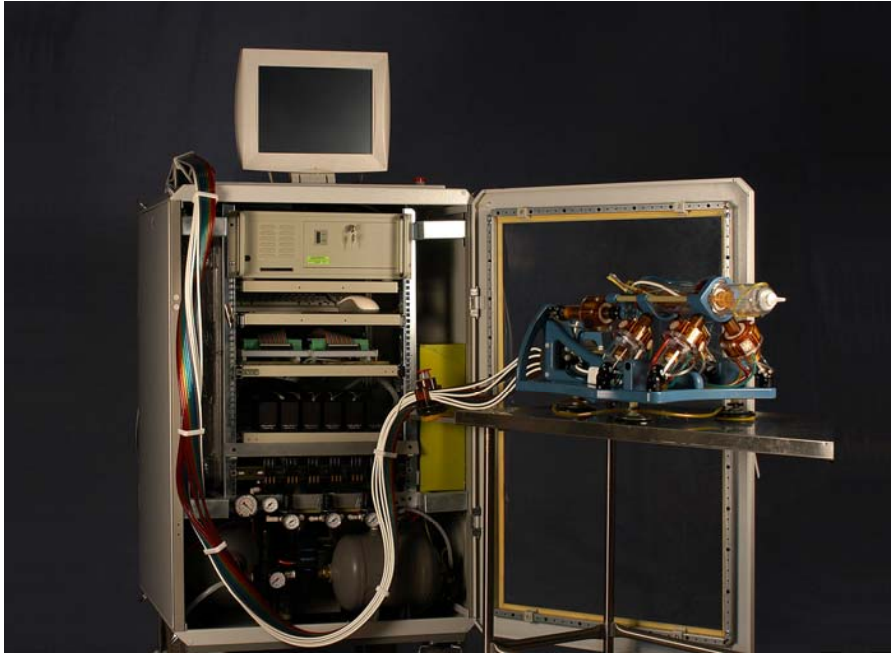

Figure 3: MrBot robot for MRI guided prostate interventions

Compared with the classic template of needle guide holes commonly used in TRUS interventions, the robot gives additional freedom of motion for better targeting. For example the skin entry point may be chosen ahead of time and targeting can be performed with angulations, which is impossible with the template, thus allowing for targeting otherwise inaccessible regions of the prostate. As such, multiple needle insertions can be performed through the same skin entry point.

The robot is controlled from a unit remotely located outside the imager's room. The robot is connected to the control cabinet by a bundle of hoses.

Precision tests in tissue mock-ups yielded a mean seed placement error of 0.72 +/- 0.36 mm [42]. With different needle drivers, the MrBot applies to various automated DIGI, such as biopsy, therapy injections, and thermal or radiofrequency ablations. The system is presently in preclinical testing with cadaver and animal experiments, but tests show very promising results and clinical trials will follow.

**C. Image-augmented remote manipulation robots:**

The combination of the two classes presented above, remote manipulation and direct image-guided robots is a very likely, highly promising direction of future developments. Augmenting guidance from medical imagers to surgical procedures could substantially improve the way that operations are being performed, and would give a clear undisputable advantage for using robotic technologies in surgery.

The NeuroArm robot under development in Canada is a good example of these technologies. Even though it may not yet operate inside the MRI scanner as planned, this may operate next to the MRI scanner and take advantage of recently acquired images to guide the surgery. This does not qualify as MRI safe and compatible, but is a "mini daVinci" with force feedback. Image processing algorithms used in robotic surgery could also improve the localization of the surgical tools and intraoperative analyses.

## V. Conclusion

The most popular medical robot thus far is perhaps the daVinci system. In only a few years since its commercial release, this technology has seen significant widespread adoption and use in operating rooms across the world. Most importantly, robotic technology such as the daVinci robot has also shown the potential of these new medical instruments and has substantially boosted the confidence of the physicians in using these technologies in the operating room environment. Concerns regarding its current capabilities and possible upgrades have been echoed, especially about its lack of force feedback. Nevertheless, daVinci represents a significant technology breakthrough and advances will improve its performance.

It is also notable that the potential of robots lies much further ahead than the accomplishments of the daVinci system. The integration of imaging with robotics holds a substantial promise, because this can accomplish tasks otherwise impossible. Image guided robots have the potential to offer a paradigm shift. The final goal of robots is to allow safer and more homogeneous outcomes with less variability of surgeon performance, as well as new tools to perform tasks based on medical transcutaneous imaging, in a less invasive way and at lower costs.

**Acknowledgments**:
The work reported from the Urology Robotics lab at Johns Hopkins has been partially supported by the National Cancer Institute (NCI) of the National Institutes of Health (NIH), the Prostate Cancer Foundation (PCF), the Patrick C. Walsh Prostate Cancer Foundation (PCW), and the Prostate Cancer Research Program of the Department of Defense (PCRP). The contents are solely the responsibility of the authors and do not necessarily represent the official views of NCI, PCF, PCW, or the PCRP.